\documentclass{article}

    \PassOptionsToPackage{numbers, compress}{natbib}
 \usepackage[preprint]{neurips_2025}

\usepackage[utf8]{inputenc} % allow utf-8 input
\usepackage[T1]{fontenc}    % use 8-bit T1 fonts
\usepackage{hyperref}       % hyperlinks
\usepackage{url}            % simple URL typesetting
\usepackage{booktabs}       % professional-quality tables
\usepackage{amsfonts}       % blackboard math symbols
\usepackage{nicefrac}       % compact symbols for 1/2, etc.
\usepackage{microtype}      % microtypography
\usepackage{xcolor}         % colors
\usepackage{amsmath}
\usepackage{tikz}
\usepackage{wrapfig}
\usetikzlibrary{arrows.meta, positioning, shapes}
\usepackage{tcolorbox}
\usepackage{tabularx}
\usepackage{enumitem}

\def\checkmark{\tikz\fill[scale=0.4](0,.35) -- (.25,0) -- (1,.7) -- (.25,.15) -- cycle;}

\title{
On the Need to Align Intent and Implementation in Uncertainty Quantification for Machine Learning
}

\author{%
  Shubhendu Trivedi\thanks{\texttt{strivedi@fnal.gov}, \texttt{shubhendu@csail.mit.edu}} \\
  \And
  Brian D. Nord\thanks{Fermi National Accelerator Laboratory, Batavia, IL 60510; Department of Astronomy and Astrophysics, \hspace*{1.8em}University of Chicago, Chicago, IL 60637; Kavli Institute for Cosmological Physics, University of Chicago \hspace*{1.8em}Chicago, IL 60637. \texttt{nord@fnal.gov}} \\
}

\begin{document}

\maketitle

\begin{abstract}
Quantifying uncertainties for machine learning (ML) models is a foundational challenge in modern data analysis. 
This challenge is compounded by at least two key aspects of the field: (a) inconsistent terminology surrounding uncertainty and estimation across disciplines, and (b) the varying technical requirements for establishing trustworthy uncertainties in diverse problem contexts. 
In this position paper, we aim to clarify the depth of these challenges by identifying these inconsistencies and articulating how different contexts impose distinct epistemic demands. 
We examine the current landscape of estimation targets (e.g., prediction, inference, simulation-based inference), uncertainty constructs (e.g., frequentist, Bayesian, fiducial), and the approaches used to map between them. 
Drawing on the literature, we highlight and explain examples of problematic mappings. 
To help address these issues, we advocate for standards that promote alignment between the \emph{intent} and \emph{implementation} of uncertainty quantification (UQ) approaches. We discuss several axes of trustworthiness that are necessary (if not sufficient) for reliable UQ in ML models, and show how these axes can inform the design and evaluation of uncertainty-aware ML systems. Our practical recommendations focus on scientific ML, offering illustrative cases and use scenarios, particularly in the context of simulation-based inference (SBI).
\end{abstract}

\section{Introduction}
\label{sec:introduction}

Uncertainty is the \emph{currency of} scientific and engineering belief, just as it is the lifeblood of operational decision--making~\citep{Savage1954-SAVTFO-2, bernardo2009bayesian, gneiting2007strictly}.  A climatologist forecasting rare events and a portfolio manager hedging macroeconomic shocks both discover, in practice, that point predictions are brittle---but calibrated doubt is actionable. Yet, the modern machine learning (ML) literature, despite its methodological breadth, offers a fragmented vocabulary~\citep{Abdar_2021} and an incoherent set of assessment tools for expressing such calibrated doubt. The issue becomes particularly acute in the ``AI for science'' literature. 
Methods like deep ensembles~\citep{lakshminarayanan2017simplescalablepredictiveuncertainty}, Bayesian neural networks~\citep{blundell2015weightuncertaintyneuralnetworks}, conformal predictors~\citep{vovk2005algorithmic}, and simulation-based inference (SBI)~\cite{cranmer2020frontier} are all used to quantify uncertainty---but often without clarifying what is being estimated, what the reported uncertainty refers to, or who it is meant to inform. 
In many cases, statistical quantities are reused in roles they were never designed to support: a variance over model outputs is treated as a stand-in for model ignorance; a prediction interval is taken as evidence about latent physical parameters. 
These moves lack what we refer to as \emph{epistemic justification}—a defensible link between the quantity being reported and the kind of claim or decision it is used to support. 
We call this phenomenon \emph{construct drift:} uncertainty is computed for one object but invoked to support conclusions about another (also see the discussion in ~\citet{box1976science} and \citet{Oberkampf_Roy_2010}). 
It is widespread, and it undermines the credibility of ML-driven science.

This paper takes the position that trustworthy uncertainty must be anchored to its inferential target, justified by its epistemic assumptions---conditions under which an uncertainty estimate can be interpreted as valid---and usable for the decisions it is meant to support. That is, there is a \textbf{need to match intent and implementation in uncertainty in scientific ML.}
We articulate a diagnostic framework that integrates three core elements: a taxonomy of estimation targets (e.g., prediction vs. parameter inference), a classification of uncertainty constructs (e.g., frequentist, Bayesian, simulation-based), and a set of trustworthiness criteria spanning theory, empirical reliability, and model correspondence.

Our contribution is organizational: we take stock of the current landscape, with an emphasis on ML for science.
Our goal is not to introduce a new method, but to offer a coherent framework for evaluating uncertainty in scientific ML. 
We begin by clarifying the estimation targets that arise in this setting and the distinct uncertainty constructs typically associated with each (\S\ref{sec:taxonomy},\S\ref{sec:uncertainty},\S\ref{sec:mapping_constructs}). 
We then articulate three axes of trustworthiness---formal guarantees, empirical reliability, and correspondence with domain structure---which serve as the basis for diagnosing alignment between uncertainty estimates and their intended role (\S\ref{sec:trust}).
These axes are used to structure a set of illustrative misalignments drawn from the literature (\S\ref{sec:misalign}) and to motivate a illustrative suite of cross-cutting diagnostics for principled evaluation (\S\ref{sec:uq_diagnostics}). 
We close with pragmatic recommendations for researchers who want their uncertainty estimates to reflect both methodological assumptions and scientific context (\S\ref{sec:recs}).

%%%%%%%%%%%%%%%%%%%%%%%%%%%%%%%%%%%%%%%%%%%%%%%%%%%%%%%%%%
\section{Taxonomy of Estimation, Inference, and Prediction}
\label{sec:taxonomy}

Before we ask how uncertainty should be quantified, we must clarify \emph{what} is being estimated. This section articulates a taxonomy of estimation targets that arise in scientific and applied ML.
Classical statistics draws a clear line between \emph{prediction}~\citep{shalev2014understanding}---estimating future, observable outcomes—and \emph{inference}~\citep{rl2002statistical, gelman2013bayesian}---recovering latent parameters believed to govern the data-generating process. But in modern ML, this boundary has become blurred. Tasks increasingly involve hybrids like \emph{predictive inference}~\citep{vovk2005algorithmic, NEURIPS2019_5103c358, foygel2021limits}, which guarantees coverage for future outcomes; \emph{indirect inference}~\citep{gourieroux1993indirect}, which estimates parameters via proxy statistics; and \emph{simulation-based inference} (SBI)~\citep{cranmer2020frontier}, which uses simulations in place of likelihoods. These are not mere terminological differences. 

Each of these targets corresponds to a different statistical object and brings with it a distinct set of expectations about what makes an estimate useful, what kind of uncertainty is appropriate, and how that uncertainty should be interpreted. 
We refer to these differing requirements as the \emph{epistemic burden} of an estimate: the specific evidentiary obligations it must satisfy to be meaningful in its intended use. 
Some estimates must guide high-stakes decisions under uncertainty; others must stand in for physical constants or encode theoretical structure; still others must remain valid despite simulator noise or model mismatch. 
For example, predicting when a battery will fail requires uncertainty estimates that support risk-aware decisions. Estimating the gravitational constant from cosmological data calls for uncertainty that enables scientific inference and theoretical consistency. These tasks differ not just in output, but in the kind of justification an estimate must carry. 

Finally, these burdens determine not only how uncertainty should be expressed, but what makes it reliable or actionable. When they are ignored or mismatched, uncertainty estimates may be precise yet \emph{epistemically invalid}---supporting claims they were never meant to justify. 

To understand this chain of reasoning, and make these distinctions concrete, we first outline six canonical estimation targets and the burdens they impose.

\begin{enumerate}[itemsep=0em, topsep=0.05em, leftmargin=1.3em]
    \item \textbf{Estimation: Summary Without Commitment.}
    Estimation, in its most generic form, refers to extracting numerical summaries from data---means, variances, low-dimensional embeddings, or transformations.   
    These may be descriptive or serve as inputs to downstream tasks, but they carry no fixed interpretive role unless embedded within a statistical model~\citep{tukey1977exploratory}.  
    That is, such summaries lack \emph{inferential semantics}—they do not, by themselves, support claims about how the data were generated or what should be believed.  
    The epistemic burden is minimal unless the estimate is used to guide decisions or justify assumptions about underlying structure.    
    \item \textbf{Prediction: Forecasting Observables.}
    The process of prediction involves a specific kind of estimation that targets observable outcomes at unobserved points---e.g., $y_{\text{new}} \sim p(y \mid x)$~\citep{shalev2014understanding}. 
    The aim is to minimize some predictive loss, such as squared error or classification loss. 
    In scientific contexts, in particular, one often needs more than a point forecast: there is a demand for uncertainty about unseen outcomes, ideally with guarantees under assumptions like exchangeability~\citep{kallenberg2005probabilistic, vovk2005algorithmic} or stationarity~\citep{kallenberg2005probabilistic}. 
    The epistemic burden here lies in operational reliability---typically understood as performance under repetition---not in terms of interpretability of model parameters or coherence with an underlying generative mechanism.
    \item \textbf{Inference: Learning Latent Parameters.} 
    Inference~\citep{rl2002statistical, gelman2013bayesian} seeks to estimate parameters $\theta$ that are not directly observable but are posited to govern the data-generating process---often physical in nature. 
    Two classical paradigms dominate: (i) A frequentist view, which treats $\theta$ as fixed and emphasizes long-run behavior under replication; (ii) A Bayesian view, which treats $\theta$ as uncertain and updates beliefs from data. 
    These frameworks share a common structure---a likelihood function---but diverge in interpretation. 
    The epistemic burden here is higher: estimates must either reflect long-run calibration or coherent belief updating. 
    In either case, inference aims to connect data with an underlying model, not just extrapolate observables.
    \item \textbf{Predictive Inference: Model-agnostic Coverage.} Predictive inference~\citep{vovk2005algorithmic, NEURIPS2019_5103c358, foygel2021limits} provides guarantees for future observations without relying on parametric assumptions or model-based reasoning. Unlike standard prediction, which may output point estimates or heuristically derived uncertainty bands, predictive inference imposes formal coverage conditions---e.g., a future observation should fall within a given region with specified probability. 
    This estimation mode is often agnostic to model correctness, focusing instead on distribution-free properties like exchangeability. 
    The epistemic burden is procedural: validity is grounded in coverage guarantees, not structural fidelity (i.e., not fidelity to a mechanistic or generative model of the data).
    \item \textbf{Indirect inference: Auxiliary Validity.} Indirect inference~\citep{gourieroux1993indirect} arises when a model’s likelihood is inaccessible but simulation is feasible. 
    Estimation proceeds by matching observed and simulated data through a tractable auxiliary model---typically one that captures salient statistical features. 
    Parameters are recovered by minimizing discrepancies in summary statistics or moment conditions. 
    Here, the epistemic burden shifts to the quality of the auxiliary model: inference is justified only insofar as the proxy summary statistics faithfully reflect the original generative process.
    \item \textbf{Simulation--based inference: : Likelihood-Free Learning.} SBI~\citep{cranmer2020frontier} generalizes indirect inference by replacing hand-crafted proxy models with flexible function approximators trained on simulated data. 
    Here, the data-generating process is defined not by a tractable likelihood but by a simulator: $\theta \mapsto \mathcal{M}(\theta) \mapsto x$. 
    From joint samples $(\theta_i, x_i)$, neural models are trained to approximate various objects of interest, such as: (i) \textbf{Neural posterior estimation} (NPE), which learns an approximation $\hat{p}(\theta \mid x)$ directly~\citep{Papamakarios2016}; (ii) \textbf{Neural likelihood estimation} (NLE), which approximates $\hat{p}(x \mid \theta)$~\citep{Papamakarios2019}; and (iii) \textbf{Neural ratio estimation} (NRE), which models the density ratio $\hat{r}(x, \theta) \propto p(x \mid \theta)/p(x)$~\citep{hermans2020likelihoodfreemcmcamortizedapproximate}. 
    These models are typically \emph{amortized}—once trained, they can produce inference results for new observations at negligible additional cost. 
    The epistemic burden here does not rest on likelihood correctness but on simulator fidelity and the generalization behavior of the learned approximator.
\end{enumerate}

As we have already discussed, each of the articulated categories reflects a different mode of reasoning, shaped by three different axes. The first axis is \emph{ontology}: what kind of entity is being posited or estimated---e.g., a future observation, a latent parameter, or the output of a simulator? 
The second axis is \emph{epistemology}: what qualifies as justification—long-run error control, internal coherence, or alignment with a mechanistic model? 
The third axis is \emph{pragmatics}: what kind of action or downstream inference the estimate is meant to support—whether it's forecasting, hypothesis testing, decision-making, or policy control? 
Together, these axes shape the inferential role that an estimate is allowed to play: this is what philosophers refer to as an estimate's \emph{epistemic warrant}~\citep{sep-transmission-justification-warrant}. 
To ignore these distinctions is to potentially mistake the kind of claim an estimate makes, and to risk applying uncertainty in ways and contexts that exceed its legitimate use.

Each member of this taxonomy pairs naturally (i.e., self-consistently) with an uncertainty construct whose logic matches the justificatory demands of the target (Table~\ref{tab:construct_mapping}). 
Prediction calls for set-valued guarantees over future observables; inference requires intervals over latent parameters, interpreted either through Bayesian posteriors or frequentist coverage; and indirect inference relies on bounds propagated through auxiliary models. 
These pairings are not arbitrary.
Applying a construct outside its intended context---say, treating a prediction set as evidence for a parameter---preserves the surface form of a guarantee, but it severs its justificatory grounding. 
The result may be numerically valid, yet epistemically misapplied.
We refer to this failure of alignment as \emph{trans-semantic transfer}: importing assurances from one inferential framework into another where their meaning no longer holds~\citep{box1976science, talts2020}. 
In the physics literature, for instance, deep ensembles are often used to report variance as a measure of “uncertainty,’’ even when that variance lacks either calibrated coverage (as in prediction) or coherent posterior semantics (as in inference). 
\emph{The slippage in interpretation exemplifies the underlying confusion in the field}, which inhibits meaningful comparison across methods, and undermines progress toward a coherent framework for uncertainty.

%%%%%%%%%%%%%%%%%%%%%%%%%%%%%%%%%%%%%%%%%%%%%%%%%%%%%%%%%%%%
\section{Uncertainty Constructs and Their Philosophical Foundations}\label{sec:uncertainty}

Having classified \emph{what} is being estimated, we next consider a complementary axis: \emph{how} uncertainty about those estimates should be represented. 
This axis is not technical alone; it is philosophical. 
Every uncertainty construct rests on a notion of what it means to be uncertain—and on what justifies that uncertainty as meaningful. 
This underlying justification is the construct’s \emph{warrant}. 
A warrant may take many forms: a long-run behavioral guarantee, a subjective degree of belief, an evidential symmetry, or an entailment between evidence and hypothesis. 
These are not interchangeable. 
They affect the kind of downstream action the uncertainty can license---whether that be operational deployment, adaptive design, hypothesis adjudication, or mechanistic explanation. 
Choosing the wrong construct can be more than a conceptual misstep. 
In scientific and engineering contexts, it can result in costly decisions: resource misallocation, premature conclusions, or spurious discoveries. 
Misalignment between what is estimated, how it is represented, and for whom it is actionable, is what we call a breakdown in the \emph{epistemic contract}.

\vspace{0.6em}
\noindent\textbf{Four canonical families.} 
Most modern uncertainty constructs fall into four broad families, each grounded in a different interpretation of probability~\citep{hacking2006emergence}, each formalizing a distinct notion of warrant.

\begin{description}[leftmargin=0.8em]
    \item[Frequentist constructs.] (Neyman, Tukey, modern error‑statistics). 
    The frequentist framework is concerned with long-run behavioral guarantees under replication. 
    For instance, confidence regions~\citep{neyman1937outline}, prediction intervals, and conformal prediction sets~\citep{vovk2005algorithmic} each guarantee long-run coverage under repeated sampling. 
    The philosophical root is Peircean fallibilism, where error control contributes to the growth of knowledge~\citep{mayo1996error}. 
    Conceptually, validity here is procedural, not propositional---it attaches to the method, not to any particular outcome~\citep{cox1958planning}.
    \item[Bayesian constructs.] (de~Finetti, Savage).
    This statistical paradigm is concerned with coherent degrees of belief updated via Bayes' rule~\citep{Savage1954-SAVTFO-2, de2017theory}. 
    The set of constructs includes credible regions and highest-posterior-density sets, which are coherent and update given a prior, yet typically lack sampling-frequency guarantees. 
    The philosophical foundation is personalism, with Dutch-book coherence serving as the constraint on rational belief~\citep{de2017theory, bovens2004bayesian}. 
    Validity here is internal: what matters is coherence of beliefs, not their long-run performance under repetition.
    \item[Fiducial and hybrid approaches.] (Fisher, Fraser, recent empirical‑Bayes). 
    These approaches treat parameters as random without invoking full priors~\citep{fraser1961fiducial, dawid2024fiducial, hannig2016generalized}. 
    The fiducial distribution is derived from the data-generating mechanism, appealing to evidential symmetry rather than prior judgment. 
    Empirical Bayes and confidence distributions fall within this pragmatist lineage. 
    The warrant here is structural: inference draws legitimacy from reversible transformations or invariant constructions rather than subjective input or asymptotic behavior.
    \item[Logical Probability.] (Keynes, Carnap).
    This is a historically significant but underrepresented tradition. 
    Here, probability is construed as a degree of entailment between evidence and hypothesis---a logical relation grounded in the information content of a proposition, not in betting behavior or physical repetition~\citep{keynes2013treatise, carnap1962logical}. 
    While not often operationalized in modern ML pipelines, this view undergirds certain maximum entropy methods, information-theoretic criteria, and recent attempts at likelihood-free inference where prior knowledge is encoded via structural constraints rather than distributional beliefs. 
    Logical probability supplies an alternate warrant: its uncertainty statements are justified if they follow from the available information, without appeal to repetition or belief.
\end{description}

These families rely on four rival interpretations of probability: behavioral frequency (von Mises), subjective degree of belief (de Finetti/Savage), evidential likelihood (Fisherian fiducial), and logical entailment (Keynes, Carnap). 
Moreover, each brings a distinct notion of epistemic warrant: performance under repetition, internal coherence, structural inversion, and evidential support, respectively. 
\emph{Our claim is not that one is superior, but that switching between them without acknowledgment—or collapsing them into a single construct—invites the construct drift identified in Section~\ref{sec:introduction}}.

Uncertainty constructs do not merely differ in style—they differ in what they mean, what assumptions they require, and what they allow us to do downstream. 
Choosing a construct implicitly commits us to a specific \emph{epistemic contract}: a logic under which uncertainty becomes justifiable and actionable. 
Ignoring this logic is what gives rise to construct drift.
A common form of this drift is what we previously called \emph{trans-semantic transfer}, where statistical guarantees from one framework (e.g., predictive coverage) are incorrectly applied within another (e.g., parameter inference), severing the link between method and meaning.

We outline below five dimensions along which constructs differ, each with implications for both modeling practice and downstream interpretation.

\begin{itemize}[itemsep=0.1em, topsep=0.05em, leftmargin=1.3em]
\item \textbf{Ontology.} Every uncertainty construct assumes a specific type of object to which it applies. This object---whether a future observable, a latent parameter, or a model component—is the construct’s \emph{ontological target}.
Using a construct on the wrong kind of object can induce a categorical misalignment. For example, treating a prediction interval (which pertains to future observables) as if it provides uncertainty over a latent parameter conflates prediction with inference—a confusion sometimes referred to as the classic “confidence vs. credibility’’ fallacy, where semantic dissonance is mistaken for statistical disagreement~\citep{jaynes2003probability}.

\item \textbf{Calibration target.} Every uncertainty construct is calibrated to a particular quantity (e.g., a future observation, latent parameter, or summary statistic for indirect inference).
Calibration means that, under certain conditions, the reported uncertainty statements (e.g., intervals or regions) are statistically valid for that object.
If uncertainty is later applied to a different object (e.g., using a prediction interval to infer a latent parameter), its guarantees no longer hold.
This disconnect can render the uncertainty irrelevant to the scientific claim or decision at hand.

\item \textbf{Computational footprint.} 
Uncertainty constructs vary widely in their computational demands.
For instance, closed-form intervals (e.g., via the Delta method or Wald approximations) are nearly free, while conformal prediction incurs \(\mathcal{O}(n)\) resampling; MCMC methods are intensive. 
Budget constraints often force a choice: practitioners must weigh not just computational cost, but also whether the chosen construct supports the epistemic burden of the decision.

\item \textbf{Decomposability.} 
Some constructs conflate all sources of uncertainty, while others distinguish between \emph{aleatoric} uncertainty (intrinsic randomness in the data) and \emph{epistemic} uncertainty (from limited knowledge or model structure).
For example, conformal prediction provides valid coverage but cannot isolate which part of the uncertainty is reducible.
In contrast, variational Bayes or hierarchical posteriors can separate these components—enabling decisions like whether further data collection will improve predictions.
This decomposability is essential in active learning, experimental design, or high-stakes domains where reducible uncertainty guides action.

\item \textbf{Robustness to model misspecification.}  
Frequentist methods may retain guarantees under mild violations. 
Bayesian posteriors can be brittle unless adjusted (e.g., via tempering).
Fiducial/hybrid and empirical-Bayes approaches offer robustness through data-driven regularization.

\end{itemize}

These distinctions motivate the necessity of deliberate pairings between estimation targets and uncertainty constructs.
A good construct is not just philosophically consistent---it enables practitioners to take action with confidence that their assumptions and inferences are aligned. Not all estimation targets enjoy equal amounts of discussion in the literature. 
The table below highlights where principled constructs are well-developed—and where conceptual or methodological gaps remain.

\begin{table}[h]
\centering
\footnotesize
\begin{tabular}{@{}lccc@{}}
\toprule
 & \multicolumn{3}{c}{\textbf{Uncertainty family}} \\
\cmidrule(lr){2-4}
\textbf{Estimation target} & Frequentist & Bayesian & Fiducial \\ \midrule
Prediction & \checkmark (conformal prediction) & – & – \\
Parameter  & \checkmark (confidence interval) & \checkmark (credible) & \checkmark$^\ast$ \\
Indirect $\psi(\theta)$ & \checkmark (approx. CI via Delta method) & – & – \\
Simulator-based $\theta$ & under-explored & \checkmark (approx. posterior via SBI) & – \\
\bottomrule
\end{tabular}
\caption{Where the literature offers, and lacks, principled constructs (illustrative). $^\ast$: limited use}
\vspace{-8mm}
\end{table}

\section{Mapping Constructs to Targets}
\label{sec:mapping_constructs}

We laid out the estimation targets (\S\ref{sec:taxonomy}) and uncertainty construct families (\S\ref{sec:uncertainty}).
Next, we align them through the lens of their epistemic warrant. 
This alignment forms what we called an \emph{epistemic contract}: a justified pathway from target, to warrant, to construct. 
This section aligns the two by asking, ``which constructs are suited to which targets, and why?'' 
Different targets entail different epistemic demands, shaping both the appropriate form of uncertainty and the kinds of decisions it licenses—whether coverage-valid, belief-coherent, error-bounded, or simulator-grounded. 
Each contract answers three questions: (i) What is the object of uncertainty: outcome, parameter, or model? (ii) What legitimizes the uncertainty statement? (iii) What kind of decisions or claims is it intended to support? 
These alignments are illustrated in Table~\ref{tab:construct_mapping}. 
Each arrow encodes a warranted inference. 
The figure is not exhaustive, but it signals which kinds of construct–target pairings are defensible.

\begin{table}[h!]
\centering
\small
\begin{tabular}{@{}l@{\hspace{0.3cm}}c@{\hspace{0.3cm}}l@{\hspace{0.3cm}}c@{\hspace{0.3cm}}l@{}}
\toprule
\textbf{Target} & & \textbf{Warrant} & & \textbf{Construct} \\
\midrule
Prediction & $\rightarrow$ & Long-run coverage (frequentist) & $\rightarrow$ & Conformal set $C_\alpha(x)$ \\
Parameter inference & $\rightarrow$ & Belief coherence (Bayesian) & $\rightarrow$ & Credible region $\mathcal{R}_\alpha$ \\
Indirect inference & $\rightarrow$ & Error rate control (frequentist) & $\rightarrow$ & Sandwich CI / Delta method \\
Simulator-based $\theta$ & $\rightarrow$ & Surrogate Bayes; learned approximation & $\rightarrow$ & NPE posterior \\
Simulator-based $\theta$ & $\rightarrow$ & Model-inversion; no prior & $\rightarrow$ & Fiducial / confidence dist. \\
Unique-event forecast & $\rightarrow$ & Evidence-to-claim strength (logical) & $\rightarrow$ & Logical support region \\
\bottomrule
\end{tabular}
\caption{Pairings between estimation targets and uncertainty constructs. 
Each carries a distinct epistemic justification. Arrows show workflow progression from target through warrant to construct.}
\label{tab:construct_mapping}
\vspace{-3mm}
\end{table}
\vspace{-2mm}

In summary, an uncertainty construct is only meaningful when its justification aligns with both the estimation target and the type of inference from which it arises. 
Each construct formalizes a distinct logic---be it long-run calibration, coherent belief, error control, or evidential support---and is valid only within the domain of that logic. 
Applying a construct outside that domain compromises both its meaning and its utility. 
The aim is not to impose rigid boundaries, but to preserve the integrity of an estimation procedure/target by ensuring that uncertainty remains anchored to its warranted role. Yet, even a construct that is logically aligned must be evaluated for trustworthiness in practice. This demands criteria beyond semantics---criteria that test whether uncertainty holds up empirically and contextually. We consider some such criteria in the next section.

%%%%%%%%%%%%%%%%%%%%%%%%%%%%%%%%%%%%%%%%%%%%%%%%%%%%%%%%%%%%
\section{Axes of Trustworthiness}
\label{sec:trust}

Trust in uncertainty estimates must be earned. In scientific applications of machine learning, this means more than quoting a variance or showing a credible interval; it requires that uncertainty constructs be tested for validity, reliability, and relevance to the scientific context.
We could consider three core axes along which uncertainty methods should be interrogated:
\vspace{-1mm}

\begin{enumerate}[itemsep=0.1em, topsep=0.05em, leftmargin=1.3em]
    \item \textbf{Formal guarantees.} Does the method offer any theoretical justification for the uncertainty it reports? For example, are there coverage results, posterior consistency theorems, or decision-theoretic bounds under well-defined assumptions? This axis concerns conditions when the construct could be valid in principle.
    \item \textbf{Empirical reliability.} Do these guarantees hold in practice? Has the method been tested via held-out simulations, posterior predictive checks, or simulation-based calibration? This axis concerns whether the construct behaves as expected when applied to data or synthetic benchmarks.
    \item \textbf{Model correspondence.}  Does the uncertainty construct reflect the structure of the domain? This includes respecting symmetries, conservation laws, known causal relationships, or measurement constraints. A construct may be statistically valid and empirically consistent, yet still misrepresent the system it is meant to model if it ignores these features.
\end{enumerate}

\begin{wrapfigure}[12]{r}{0.53\textwidth}
% note by ST the 12 here is the number of lines the figure wraps aorund. 
\centering
\vspace{-10pt}
\begin{tikzpicture}[
  scale=0.65,
  font=\small,
  mainbox/.style={
    draw, rounded corners=2pt, minimum width=3.2cm, minimum height=1cm,
    thick, align=center, fill=#1!15
  },
  centerbox/.style={
    draw, rounded corners=3pt, thick, fill=white, align=center,
    minimum width=3.5cm, minimum height=1.1cm
  },
  egtext/.style={
    font=\scriptsize, color=gray!190, align=center
  }
]

% Define coordinates
\coordinate (A) at (0, 4);     % Formal Guarantees
\coordinate (B) at (6, 4);     % Empirical Reliability
\coordinate (C) at (3, 0);     % Model Correspondence
\coordinate (Center) at (3, 2.2); % Trustworthy Uncertainty

% Triangle edges
\draw[thick, densely dotted, gray!80] (A) -- (B);
\draw[thick, densely dotted, gray!80] (A) -- (C);
\draw[thick, densely dotted, gray!80] (B) -- (C);

% Arrows to center
\draw[thick, densely dotted, gray!80] (A) -- (Center);
\draw[thick, densely dotted, gray!80] (B) -- (Center);
\draw[thick, densely dotted, gray!80] (C) -- (Center);

% Nodes
\node[mainbox=blue, anchor=south] at (A) {\textbf{Formal Guarantees}};
\node[mainbox=green, anchor=south] at (B) {\textbf{Empirical Reliability}};
\node[mainbox=orange, anchor=north] at (C) {\textbf{Model Correspondence}};
\node[centerbox] at (Center) {\textbf{Trustworthy} \\ \textbf{Uncertainty}};

% e.g. lines
\node[egtext, below=1.5pt of A] {
\textit{e.g.} conformal coverage,\\
posterior bounds
};

\node[egtext, below=1.5pt of B] {
\textit{e.g.} SBC, predictive\\
checks
};

\node[egtext, above=1.5pt of C] {
\textit{e.g.} simulator fidelity,\\
domain constraints
};

\end{tikzpicture}
\vspace{-5pt}
\end{wrapfigure}

These axes are not interchangeable. A construct may satisfy long-run coverage (axis 1) and benchmark well on synthetic data (axis 2), yet remain epistemically meaningless if it ignores how the physical system works (axis 3). 
Scientific UQ is not just about performance---it is about warrant. 
To claim that a model knows what it does not know, uncertainty must hold up across all three axes.

\textbf{Operationalizing the Axes: The Scientific SBI Checklist.} To illustrate these axes, we consider the case of simulation-based inference (SBI), where these axes manifest as a series of necessary challenges---each probing a different failure mode. We present them not as best practices, but as minimum conditions for accountability:

\begin{enumerate}[itemsep=0.1em, topsep=0.05em, leftmargin=1.3em]
\item \textbf{Theory check (guarantee).} What does the method claim to get right? This includes posterior approximation guarantees (e.g., amortization bounds in NPE), or coverage results under exchangeability (as in conformal methods). 
\item \textbf{Forward checks (data space).} Can the posterior, when sampled through the simulator, reproduce the distribution of observed data? This is not just a visual check. It tests whether the uncertainty respects the observable footprint of the phenomenon.
\item \textbf{Inverse checks (parameter space).} Can we recover known parameters from simulated data? Does the posterior place appropriate weight on ground-truth values, or does it systematically under- or over-cover? This tests identifiability and calibration.
\item \textbf{Degeneracy mapping.} In hierarchical models or high-dimensional simulators, are there directions in parameter space that are observationally equivalent? Have these been identified, visualized, or tested? Without this, posteriors can be misleadingly sharp or diffuse.
\item \textbf{Global structure comprehension.} Has the joint distribution \( p(\theta, x) \) been examined in full? This includes characterizing regions of the prior that produce nonsensical outputs, mapping simulator failure regimes, and understanding data manifold structure. This is not auxiliary---it is necessary situational awareness.

\end{enumerate}

Together, these checks instantiate a principle: \emph{\textbf{uncertainty is not a property of a method, but of a method-in-some-context}}. Trustworthy UQ arises when inference is embedded in a loop of adversarial self-testing---where every uncertainty claim is a provisional hypothesis, subject to refutation by forward and inverse challenge.

%Procedurally, I was also thinking that most SBI models in the physical sciences need the the following:

%\begin{enumerate}
%    \item A guarantee (a theory check)
%    \item Inverse checks (related to parameter space checks) 
%    \item Forward checks (related to data space checks)
%    \item Parameter space checks
%    \item Data space checks
%    \item Full data and parameter space overview: for simulations, having full (mathematical, quantitative, and visualization) of the model. Basically looking at all the patterns in the data and understanding degeneracies between parameters, such as in hierarchical model. This is similar to Karpathy's advice about becoming one with the data. physicists still don't do this systematically. 
%\end{enumerate}

%%%%%%%%%%%%%%%%%%%%%%%%%%%%%%%%%%%%%%%%%%%%%%%%%%%%%%%%%%%%

\section{Misaligned Uncertainty in Scientific ML}
\label{sec:misalign}

ML is increasingly embedded in scientific workflows—used to emulate simulations, infer latent structure, and predict physical properties. Yet, the uncertainty constructs adopted in these pipelines are often unexamined, and rarely scrutinized in terms of their scientific function. 
In many cases, these constructs serve as placeholders for ``interpretability'', lacking clear justification for the claims they support or the decisions they are meant to guide.
We highlight some general patterns of misalignment.

\noindent
\textbf{Semantic misalignment: epistemic $\neq$ systematic.} (violates Axes 1 \& 3).
In physics, uncertainties are often categorized as either \emph{statistical} (variation due to limited data) or \emph{systematic} (persistent bias from instruments, calibration, or imperfect theory)~\citep{oberkampf2004verification}. 
In ML, by contrast, the dominant division is \emph{aleatoric} (irreducible noise) versus \emph{epistemic}  (model or data-driven uncertainty)~\citep{hullermeier2021aleatoric}\footnote{The notions of aleatoric and epistemic uncertainties are not absolute---they are context dependent. A change in context could change one into the other~\citep{hullermeier2021aleatoric, der2009aleatory}.}.These vocabularies are not interchangeable. 
For instance, calibration errors in physical measurements are systematic, but are often modeled as aleatoric noise in ML frameworks. Similarly, model misspecification in SBI---central in cosmology or climate science---is rarely captured by epistemic constructs unless explicitly addressed. This conflation leads to interpretive errors: a Bayesian neural network’s variance may be taken to represent physical uncertainty, when in fact it encodes posterior dispersion over model parameters, not over nature. Such slippages can have material consequences: misprioritized experiments, overconfident forecasts, or misleading estimates of scientific risk.

\textbf{Scientific decisions require target-consistent uncertainty.} (violates Axes 2 \& 3).
A 95\% confidence interval on cosmological parameters may be useful for statistical assessment—but it does not indicate where to point a telescope. Conversely, a conformal interval around a galaxy cluster’s predicted mass may guarantee valid coverage, but folds in multiple noise sources and omits priors or instrumental models. Each construct is valid within its own semantics but may be misleading when transferred into workflows without regard for its intended target or action. In battery lifetime modeling, prediction intervals may calibrate on held-out data but ignore drift in operational use. In epidemiological forecasting, coverage guarantees may hold for historical data but break under interventions or regime changes. Scientific actions require uncertainty that is not just valid, but inferentially appropriate.

\textbf{Method-first papers obscure uncertainty usage.} (violates Axes 1, 2, \& 3). 
A growing genre of ML for science papers focus on using new ML architectures for different applications and report uncertainty---e.g., normalizing flows for cosmological parameter estimation, invertible networks for chemical structure inference, or graph neural nets for materials screening---without critically assessing whether the uncertainty produced by these models has a meaningful interpretation in the scientific context.
Uncertainty is reported, often in the form of standard deviation or entropy over predictive samples, but the constructs are often misaligned; not diagnosed or validated properly, or connected to scientific decision-making endpoints.
In SBI, this problem is acute. 
The posterior from a neural density estimator (e.g., NPE or NRE) may appear calibrated on test simulations but carries no epistemic status unless it reflects both simulator fidelity and model misspecification. 
Most SBI papers do not check this. 
Posterior predictive checks are rare; sensitivity to simulator parameters is usually ignored. 
The complexity of the model is used, sometimes, to focus on just the predictive output because it is \emph{learned}.

To add to the above, below we give some more concrete examples---common in the literature---where uncertainty constructs, though valid within their own framework, are applied outside their inferential scope---exceeding or distorting their epistemic warrant---resulting in misleading or incoherent scientific conclusions.

\textbf{Variance \emph{versus} uncertainty} (violates Axes 1 \& 2). In molecular-property prediction, deep ensembles often report variance across models as ``epistemic uncertainty.'' Yet, this variance carries no formal calibration guarantee and is rarely tested for frequentist coverage or Bayesian coherence. It is neither necessary nor sufficient for either type of statistical warrant. \citet{yang2023explainable} move beyond this heuristic by introducing atom-level decompositions that disentangle aleatoric and epistemic components more rigorously, producing uncertainty estimates that are better grounded. The same misalignment has also been reported in astrophysics. See \citet{loredo2012bayesian} for some examples. 

\textbf{Coverage Without Ontological Alignment} (violates Axes 2 \& 3). In astrophysical spectral energy distribution (SED) modeling, conformalized quantile regression has been employed to generate prediction intervals with nominal coverage guarantees. However, these intervals often fail to account for the hierarchical and generative structures inherent in astrophysical data, such as the relationships between stellar mass, dust content, and star formation rates. By treating observations as exchangeable and not incorporating domain-specific knowledge, the resulting uncertainty estimates may conflate different sources of variability, leading to intervals that lack interpretability and fail to inform scientific inquiry effectively. This highlights the necessity of aligning statistical methods with the underlying scientific ontology to ensure that uncertainty quantification is both reliable and meaningful.

\textbf{Inference without simulator fidelity} (violates Axes 1 \& 3).  Neural ratio estimation (NRE) is a popular likelihood-free method in collider physics. When trained on fast surrogate simulators, NRE can outperform ABC benchmarks---but inherits untracked bias. \citet{delaunoy2022towards} show that standard NRE often yields overconfident posteriors with poor frequentist coverage unless explicitly regularized. This issue is amplified when high-fidelity simulators are reinstated, or when surrogate assumptions break down, leaving the posterior without coverage and without simulator-based warrant.

\textbf{Warrant Confusion Across Paradigms} (violates Axes 1 \& 3).
In high-energy physics, it's common to report both frequentist p-values and Bayesian credible regions for the same signal detection task~\citep{Lyons_2013}. These constructs rely on fundamentally different notions of justification---long-run error rates versus conditional belief coherence. When diffuse priors make Bayesian intervals narrower, practitioners sometimes treat this as a contradiction rather than a difference in warrant. This confusion stems from mixing constructs without clarifying their inferential semantics.

\textbf{Calibration Without Causal Insight} (violates Axes 2 \& 3).
In clinical-based predictions, conformal predictors can achieve marginal coverage across the full population. Yet they often fail to explain why uncertainty expands for particular subgroups---e.g., nephrology patients~\citep{lin2022conformal, lin2022conformal2}. This failure arises because conformal constructs calibrate for coverage but not for explanation. Without hierarchical structure or causal modeling, the construct is warranted for population-level prediction---but not for stratified insight or intervention.

\textbf{Toward a scientific standard.}
The examples above reinforce a central point: uncertainty constructs are only as trustworthy as the epistemic contract they satisfy. This contract must span all three axes: (1) formal guarantees that define what the construct means; (2) empirical reliability that tests whether it holds in practice; and (3) correspondence with domain structure that ensures it remains grounded in scientific context. A scientifically valid uncertainty estimate must be traceable to its source, justified by its logic, and usable for the decision it claims to support. This requires treating UQ not as an afterthought, but as a core object of analysis. It translates to demanding transparency (“what does this interval mean?”), robustness (“does it generalize?”), and alignment (“what action does it guide?”). While tools like simulation-based calibration (SBC)~\citep{talts2020validatingbayesianinferencealgorithms} provide partial answers in SBI, no consensus yet exists on whether they are sufficient. We argue that the next phase of scientific ML requires principled scrutiny of uncertainty constructs, not just model performance.

\textbf{Cross‐cutting Diagnostics for Uncertainty Constructs.}
\label{sec:uq_diagnostics} 
To illustrate how uncertainty constructs can be interrogated across different dimensions, whether philosophical, statistical, or computational, we provide an illustrative list of cross-cutting diagnostics in Table~\ref{tab:uq_diagnostics}. 
These checks are not exhaustive, nor universally applicable, but they show how different properties can be probed systematically. The aim is not to prescribe a fixed protocol, but to encourage deliberate testing aligned with the construct’s intended use, core assumptions, and inferential role.

%A scientifically valid uncertainty construct must be traceable to its source, justified by its inferential logic, and usable for the decisions it claims to support. This requires treating UQ not as an afterthought, but as a core object of analysis. Transparency is the first step: What does a reported interval mean? What assumptions does it require? What quantity is being conditioned on? The second step is diagnosis: Is the construct calibrated? Robust? Aligned with the action it is meant to inform? While tools like simulation-based calibration (SBC) provide partial answers in SBI, no consensus yet exists on whether they are sufficient. We argue that the next phase of scientific ML requires principled scrutiny of uncertainty constructs, not just model performance.

%%%%%%%%%%%%%%%%%%%%%%%%%%%%%%%%%%%%%%%%%%%%%%%%%%%%%%%%%%%%
\section{Recommendations and Outlook}
\label{sec:recs}

The flexibility of modern ML tools, especially simulation-based inference (SBI), has expanded the space of scientific modeling. But that flexibility also makes it easier for uncertainty constructs to drift away from their intended use. What’s needed is not just methodological innovation, but principled constraint: models must be evaluated in context, uncertainty estimates interrogated for their validity, and methods aligned with decision goals. Below, we outline actionable recommendations, drawn from the preceding framework, that can guide the responsible use of uncertainty in scientific ML.

\begin{itemize}[itemsep=0.1em, topsep=0.5em, leftmargin=1.5em]

\item \textbf{Declare the inference chain.} Make explicit (i) the estimation target (e.g., future battery failure time; cosmological parameter), (ii) the loss or decision goal (e.g., minimize downtime risk; test theoretical model fit), (iii) the uncertainty construct used  (e.g., conformal prediction set; Bayesian credible interval), and (iv) the warrant it is meant to carry  (e.g., marginal coverage; posterior belief coherence).

\item \textbf{Match construct to context.} Choose constructs based on what the uncertainty is meant to support—e.g., exploration, control, forecasting, hypothesis testing—not based on availability or familiarity.

\item \textbf{Check both forward and inverse validity.} Validate uncertainty both in data space (e.g., posterior predictive checks~\citep{rubin1984bayesianly, gelman1996posterior}, calibration curves~\cite{niculescu2005predicting, degroot1983comparison}) and parameter space (e.g., simulation-based calibration~\citep{talts2020validatingbayesianinferencealgorithms}, inverse coverage).

\item \textbf{Avoid conflating constructs.} A prediction interval is not a credible region; a standard deviation over predictions is not epistemic uncertainty. Misuse of these items leads directly to construct drift, and thus invites misinterpretation.

\item \textbf{Engineer evaluation to the decision.} For example, if the scientific decision hinges on false-negative rates, prioritize stratified calibration (e.g. \citep{gibbs2024conformalpredictionconditionalguarantees}); if in a physical problem, phase boundaries matter, stratify coverage by regime (e.g., as in \citep{PhysRevMaterials.7.025201})

\item \textbf{Use the simulator as an instrument.} Where available, use simulators not just to train, but to test: perturb parameters, introduce plausible misspecification, and measure UQ behavior under controlled change.

\item \textbf{Benchmark beyond IID.} Coverage and calibration under i.i.d. conditions are insufficient. Examine domain shifts, covariate drift, or structured noise in test regimes.

\item \textbf{Study model stability.} Examine sensitivity of results to dataset realizations, initialization seeds, or retraining. Stability is a proxy for epistemic trustworthiness.

\item \textbf{Invest in cross-disciplinary clarity.} When borrowing terms like “epistemic,” “systematic,” or “confidence,” define them with respect to both their statistical and scientific context.

\end{itemize}

Improving UQ in scientific ML does not require solving foundational questions in the philosophy of science. But it does require practical discipline: defining what is being estimated, justifying the uncertainty attached to it, and validating whether that uncertainty supports the decisions or claims it is meant to inform. We advocate for this kind of \emph{epistemic hygiene}--—not as a constraint, but as an enabling structure. The payoff is not just cleaner semantics, but more decisive modeling. By aligning estimation targets with uncertainty constructs and validation tools, we enable models to play a trustworthy role in scientific inference---making uncertainty a vehicle for insight, not confusion, and supporting the kind of transparent, cumulative progress that scientific ML now urgently demands.

%%%%%%%%%%%%%%%%%%%%%%%%%%%%%%%%%%%%%%%%%%%%%%%%%%%%%

%%%%%%%%%%%%%%%%%%%%%%%%%%%%%%%%%%%%%%%%%%%%%%%%%%%%%%%%%%%%
% \begin{ack}
% Use unnumbered first level headings for the acknowledgments. All acknowledgments
% go at the end of the paper before the list of references. Moreover, you are required to declare
% funding (financial activities supporting the submitted work) and competing interests (related financial activities outside the submitted work).
% More information about this disclosure can be found at: \url{https://neurips.cc/Conferences/2025/PaperInformation/FundingDisclosure}.

% Do {\bf not} include this section in the anonymized submission, only in the final paper. You can use the \texttt{ack} environment provided in the style file to automatically hide this section in the anonymized submission.
% \end{ack}

\newpage

\bibliographystyle{unsrtnat}
\bibliography{references}

%%%%%%%%%%%%%%%%%%%%%%%%%%%%%%%%%%%%%%%%%%%%%%%%%%%%%%%%%%%%

\newpage
\appendix

\section{Cross-Cutting Diagnostics}

\begin{table}[h!]
\centering
\renewcommand{\arraystretch}{1.35}
\scalebox{0.85}{
\begin{tabular}{@{}p{3.8cm}p{5.0cm}p{5.2cm}@{}}
\toprule
\textbf{Diagnostic} & \textbf{What to Measure} & \textbf{Example Tools or Techniques} \\
\midrule
\multicolumn{3}{l}{\textbf{Formal Validity}} \\
\midrule
\textbf{Marginal coverage} & Frequency with which intervals contain true values across held-out samples & SBC (Simulation-Based Calibration), coverage plots, PIT (Probability Integral Transform) histograms, jackknife+ \\
\textbf{Construct provenance} & Theoretical justification for the uncertainty statement & Formal derivation, bibliographic lineage \\
\midrule
\multicolumn{3}{l}{\textbf{Empirical Reliability}} \\
\midrule
\textbf{Conditional coverage} & Calibration within subgroups or across covariates & Groupwise conformal calibration, conditional coverage curves \\
\textbf{Interval sharpness} & Expected size (length/volume) of predictive sets relative to coverage & Width–coverage tradeoff, log-length plots \\
\textbf{Out-of-distribution behavior} & Predictive confidence under distribution shift or rare inputs & OOD scoring, entropy spikes, tail credibility plots \\
\textbf{Coverage convergence} & Improvement of calibration with more data & Coverage learning curves, repeated subsampling \\
\midrule
\multicolumn{3}{l}{\textbf{Model Correspondence}} \\
\midrule
\textbf{Aleatoric/epistemic separation} & Ability to decompose total uncertainty into noise vs. knowledge components & Deep ensemble + conformal, dropout variance vs. mean spread \\
\textbf{Construct consistency} & Agreement across models or inference algorithms & Cross-method comparison, posterior overlap measures \\
\textbf{Composability} & Does UQ remain valid when models are chained or modularized? & Simulator composition tests, plug-in error propagation \\
\textbf{Structural priors} & Does uncertainty respect known structure (e.g., symmetry, conservation)? & Group-equivariant priors, physics-informed nets \\
\midrule
\multicolumn{3}{l}{\textbf{Decision Alignment*}} \\
\midrule
\textbf{Task-aware utility} & Is uncertainty informative for decisions (e.g., abstention, ranking)? & Utility-aware calibration, selective prediction loss curves \\
\textbf{Counterfactual consistency} & Does uncertainty remain stable under plausible interventions? & Sensitivity analysis, policy-driven stress tests \\
\textbf{Construct declaration} & Explicit documentation of what the construct assumes and supports & “Uncertainty cards,” epistemic audit logs, type legends in figures \\
\bottomrule
\end{tabular}
}
\caption{
Cross-cutting diagnostics for evaluating uncertainty constructs. The table is illustrative. Each construct should be tested along dimensions that reflect its intended use, assumptions, and decision context. Categories align with the three axes of trustworthiness introduced in Section~\ref{sec:trust}, with an additional axis (*) for decision alignment, critical in scientific deployments.
}
\label{tab:uq_diagnostics}
\end{table}

\end{document}